\documentclass[10pt,twocolumn,letterpaper]{article}

\usepackage[pagenumbers]{cvpr} 

\usepackage{graphicx}
\usepackage{amsmath}
\usepackage{amssymb}
\usepackage{booktabs}
\usepackage{multirow}
\usepackage[dvipsnames,table,xcdraw]{xcolor}
\usepackage{color, colortbl}
\definecolor{Gray}{gray}{.9}
\newcolumntype{g}{>{\columncolor{Gray}}c}

\usepackage[pagebackref,breaklinks,colorlinks]{hyperref}

\usepackage[capitalize]{cleveref}
\crefname{section}{Sec.}{Secs.}
\Crefname{section}{Section}{Sections}
\Crefname{table}{Table}{Tables}
\crefname{table}{Tab.}{Tabs.}

\newcommand{\mysection}[1]{\vspace{2pt}\noindent\textbf{#1}}

\newcommand{\studentNetwork}{\mathbf{S}}
\newcommand{\teacherNetwork}{\mathbf{T}}
\newcommand{\studentNetworkCopy}{\mathbf{S}_c}

\newcommand{\video}{\mathcal{V}}
\newcommand{\loss}{\mathcal{L}}
\newcommand{\distance}{L}
\newcommand{\probability}{p}
\newcommand{\image}{x}
\newcommand{\imageset}{X}
\newcommand{\rate}{r}
\newcommand{\task}{\mathcal{T}}
\newcommand{\prediction}{\hat{y}}
\newcommand{\pseudogroundtruth}{\tilde{y}}
\newcommand{\pseudogroundtruthset}{\tilde{Y}}
\newcommand{\importance}{\mathcal{I}}
\newcommand{\domain}[1]{D^{#1}}
\newcommand{\videodomain}[1]{\video^{#1}}
\newcommand{\metric}{\mathcal{M}}
\newcommand{\metricsegmentation}{mIoU}

\newcommand{\updateFunction}{f_U}
\newcommand{\selectionFunction}{f_S}

\newcommand{\onlineDataset}{\mathcal{D}}
\newcommand{\ODSize}{N}
\newcommand{\RBSize}{M}

\newcommand{\weight}{\theta}

\newcommand{\regularisation}{\mathcal{R}}

\newcommand{\evaluationsize}{I}

\newcommand{\updateFreq}{k}

\definecolor{darkRed}{rgb}{0.9, 0.17, 0.31}

\def\cvprPaperID{10} 
\def\confName{CVPRW}
\def\confYear{2023}
\newcommand\blfootnote[1]{%
  \begingroup
  \renewcommand\thefootnote{}\footnote{#1}%
  \addtocounter{footnote}{-1}%
  \endgroup
}

\begin{document}

\title{Online Distillation with Continual Learning for Cyclic Domain Shifts}

\author{Joachim Houyon$^{1,*}$ 
\and
Anthony Cioppa$^{1,2,*}$  
\and
Yasir Ghunaim$^2$  
\and
Motasem Alfarra$^2$ 
\and
Anaïs Halin$^1$  
\and
Maxim Henry$^1$  
\and
Bernard Ghanem$^2$  
\and
Marc Van Droogenbroeck$^1$  
\and
$^1$ University of Liège\quad $^2$ KAUST
}
\maketitle

\begin{abstract}
In recent years, online distillation has emerged as a powerful technique for adapting real-time deep neural networks on the fly using a slow, but accurate teacher model. 
However, a major challenge in online distillation is catastrophic forgetting when the domain shifts, which occurs when the student model is updated with data from the new domain and forgets previously learned knowledge.
 In this paper, we propose a solution to this issue by leveraging the power of continual learning methods to reduce the impact of domain shifts.
Specifically, we integrate several state-of-the-art continual learning methods in the context of online distillation and demonstrate their effectiveness in reducing catastrophic forgetting.
Furthermore, we provide a detailed analysis of our proposed solution in the case of cyclic domain shifts.
Our experimental results demonstrate the efficacy of our approach in improving the robustness and accuracy of online distillation, with potential applications in domains such as video surveillance or autonomous driving. Overall, our work represents an important step forward in the field of online distillation and continual learning, with the potential to significantly impact real-world applications.
\blfootnote{\textbf{(*)} Equal contributions \\Contacts: joachim.jouyon@student.uliege.be, anthony.cioppa@uliege.be. 
Data/code available at \href{https://github.com/Houyon/online-distillation-cl}{github.com/Houyon/online-distillation-cl}.
}
\end{abstract}

\section{Introduction}
\label{sec:intro}

\begin{figure}
    \centering
    \includegraphics[width=0.9\linewidth]{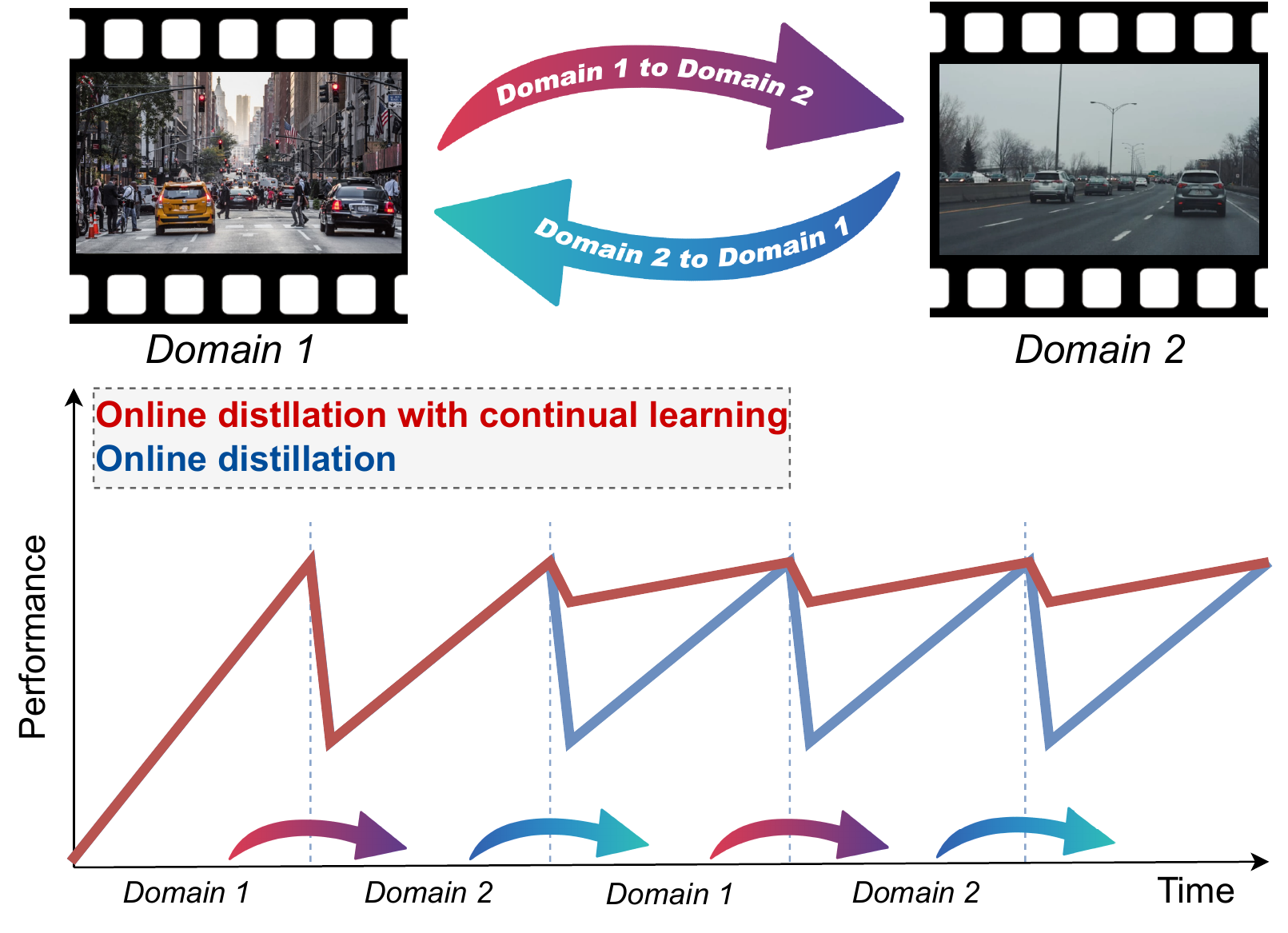}
    \caption{\textbf{Online distillation with continual learning.} When cyclic domain shifts occur in long videos, the online distillation framework proposed by Cioppa~\etal~\cite{Cioppa2019ARTHuS} forgets the previously acquired knowledge as it fine-tunes on the current domain. In this work, we study the inclusion of state-of-the-art continual learning methods inside the online distillation framework to mitigate this catastrophic forgetting around the domain shifts.}
    \label{fig:graphical}
\end{figure}

Deep Neural Networks (DNNs) have shown remarkable performance on various computer vision tasks thanks in part to the assumption that the training and testing data are identically distributed~\cite{he2016resnet, liang2015rcnn, schmidhuber2015deep}. 
However, DNNs' performance degrade significantly when tested on out-of-distribution data, such as testing data that contains domain shifts relative to the training data~\cite{imagenetc,3dcc}.
Even worse, DNNs tend to forget previously learned distributions when learning continually on a stream of tasks~\cite{kirkpatrick2017overcoming}.
This performance loss is a major concern because domain shifts are likely to occur in real-world deployments due to changes in brightness between day and night, weather conditions across seasons, and sensor perturbations~\cite{sakaridis2021acdc}. 
Therefore, it is essential to develop algorithms that can enable DNNs to adapt to such domain shifts and maintain high performance in real-world settings.

Continual learning aims at building machine learning models that can learn from a continuous stream of data without forgetting previously learned knowledge~\cite{li2017learning, chaudhry2019continual, He2020Incremental}.
We investigate a practical scenario of online continual learning~\cite{cai2021online}. 
Specifically, we consider cyclic domain shifts where a stream of data consistently alternates in revealing new \emph{unlabeled} data from one of two distributions for a period of time. 
For instance, consider an autonomous driving system that frequently travels between cities and countrysides, where the distribution of instances varies between the two scenes. 
Such domain variation can cause the online learner to fail in adapting to this distribution shift, raising concerns about the real-world deployment of such systems. 
While online continual learning has been studied in several contexts, such as domain incremental learning~\cite{Garg2022MultiDomain}, unsupervised domain adaptation~\cite{sun2016deep}, and test-time adaptation~\cite{tent}, these works typically analyze the more general, and potentially less realistic, setup where domain variations are unconditional. 
Our focus on cyclic domain shifts enables us to explore a pragmatic setting and develop novel algorithms that can better adapt to these changes.

In this work, we propose a novel approach to address the challenge of adapting to cyclic domain shifts in the context of online domain incremental learning. 
Specifically, we employ a previously published real-time online distillation technique~\cite{Cioppa2019ARTHuS} to learn from the unlabeled cyclic stream of data. 
Online distillation asynchronously updates a student-teacher based approach on the received data, which enables the model to continually learn from new data.
However, we found that the cyclic domain shift can cause the student to forget the previously learned domain, leading to a significant loss in performance. 
To mitigate this undesirable effect, we combine online distillation with state-of-the-art continual learning as shown in Figure~\ref{fig:graphical}, leveraging both regularization- and replay-based approaches from the continual learning literature. 
Our proposed approach enables the student to effectively adapt to cyclic domain shifts and maintain high performance over time, making it suitable for real-world deployment.

\mysection{Contributions.} We summarize our contributions in two points:
\textbf{(i)} We define the cyclic online continual learning problem setup and propose corresponding evaluation metrics.
\textbf{(ii)} We combine online distillation with both regularization- and replay-based continual learning approaches to better learn on cyclic domains. We conduct experiments on the proposed stream where we show that our approach mitigates the forgetting of the original online distillation framework.



\section{Related Work}
\label{sec:relatedwork}


\mysection{Domain shifts.}
A domain shift is a change in the statistical distribution of data between different domains~\cite{Farahani2021ABrief}.
This phenomenon is commonly observed at test time in open-world scenarios~\cite{Lim2023TTN,Wang2022Continual,Azimi2022Selfsupervised, Ghunaim2023RealTime-arxiv}.
In autonomous driving, domain shift can be caused by many diverse factors~\cite{Sun2022SHIFT}, such as different environments (\eg, rural or urban roads), lighting conditions (\eg, day or night), weather conditions (\eg, sunny or snowy)~\cite{sakaridis2021acdc}, traffic conditions or even differences in the appearance of roads or traffic signs across different countries~\cite{Yu2020BDD100K}.
However, it is crucial for autonomous vehicles to have algorithms that are robust to these dynamic domain shifts in order to constantly be able to perceive and understand their surrounding environment to avoid obstacles.
Domain adaptation is an active area of research that aims at addressing the domain shift problem, especially in open-world applications such as autonomous vehicles~\cite{Li2023Domain,Pierard2023Mixture,Panagiotakopoulos2022Online,Sun2022SHIFT,Luo2019Taking}, where data is collected in a highly dynamic environment.
In this work, we study the particular case of cyclic domain shifts in the field of autonomous driving, where the domains can be represented as a succession of \textit{highway} and \textit{downtown} driving conditions.

\mysection{Online distillation.}
In the field of deep neural networks, there is a trade-off between speed, performance, and generalizability across multiple domains. 
While the best-performing models often exhibit high performance across diverse domains, they tend to be memory-greedy for embedded systems or too slow for use in real-time applications~\cite{Xie2021SegFormer,Zhao2017Pyramid,Zheng2021Rethinking}. 
In contrast, lightweight and fast networks show good performance on smaller domains but lack generalizability~\cite{Cioppa2018ABottomUp}. To address this issue, Cioppa~\etal~\cite{Cioppa2019ARTHuS} proposed an online distillation approach for videos, that enables the online training of a lightweight student network using a slower, larger teacher model. At test time, the teacher provides pseudo ground truths to the student, allowing it to specialize in the specific domain being analyzed. The student model therefore adapts to changing video conditions, even matching the performance of the slower teacher. This online distillation approach may be used for different tasks such as semantic segmentation~\cite{Cioppa2019ARTHuS} or multi-modal object detection~\cite{Cioppa2020Multimodal}. However, this technique experiences a temporary loss of performance during domain shifts. 
In this paper, we investigate several continual techniques to mitigate the effects of catastrophic forgetting in online distillation, particularly in cases of cyclic domain shifts.
We combine online distillation with both regularization- and replay-based approaches for a better continual learning scheme.

\mysection{Continual learning.} 
Continual Learning~(CL) aims at learning from data arriving as a stream with changing distribution~\cite{mccloskey1989catastrophic, french1999catastrophic}.
However, this learning paradigm face the catastrophic forgetting challenge, that is, previously learned knowledge is forgotten when adapting to the newly arriving data samples~\cite{kirkpatrick2017overcoming,chaudhry2019continual}.
One approach of mitigating the forgetting effect is regularizing the training process through constraining the changes of important network parameters~\cite{kirkpatrick2017overcoming, aljundi2018memory, chaudhry2018riemannian} or performing knowledge distillation~\cite{li2017learning, smith2021always, gao2022rdfcil}.
Alternatively, replay-based methods rehearse previously seen examples by storing a subset of the observed data in a replay buffer~\cite{chaudhry2019continual, lopez2017gradient, aljundi2019online, Prabhu2020GDumb}.
While both approaches were originally proposed for class-incremental setup and classification task, they were recently extended to the more realistic domain incremental setup and the more challenging semantic segmentation task~\cite{Alfarra2022SimCS-arxiv,Garg2022MultiDomain}.
Nevertheless, prior art assumes fully supervised setups where the stream reveals labeled data for the student learner.
To that end, we analyze the domain incremental setup for semantic segmentation under an unsupervised setup.

\section{Methodology}
\label{sec:method}

\begin{figure*}
    \centering
    \includegraphics[width=\linewidth]{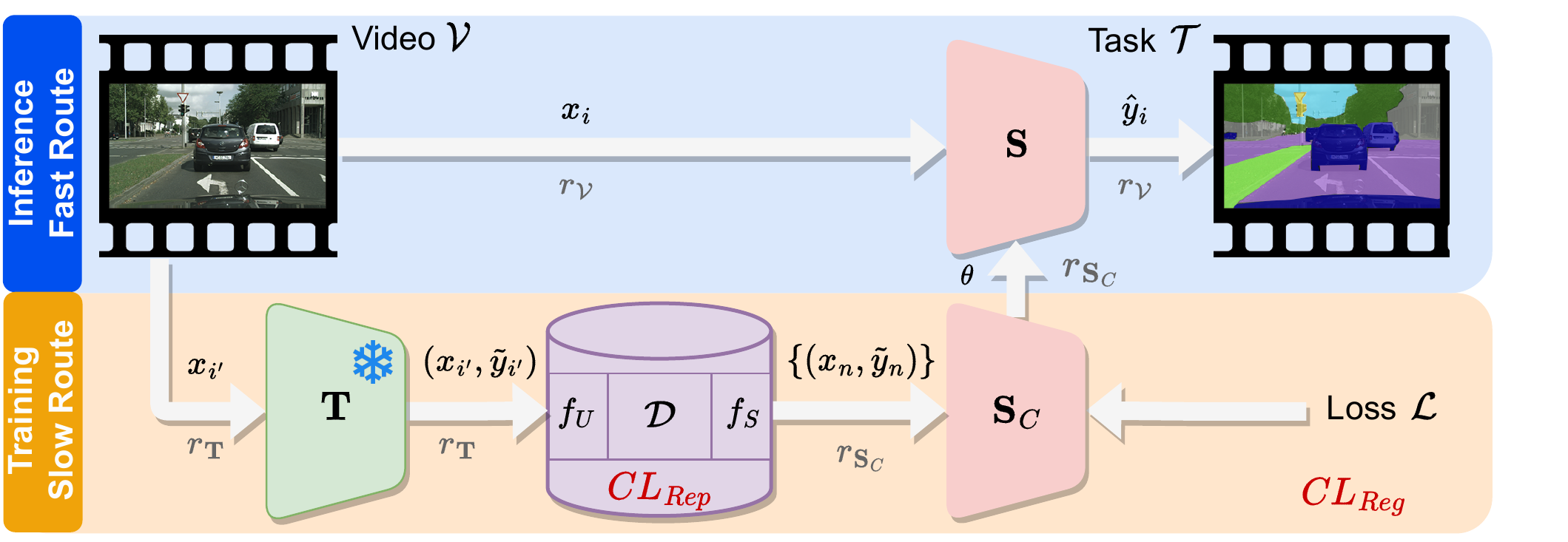}
    \caption{\textbf{Online distillation}. The framework is composed of a fast and a slow route. In the fast route (inference), the video stream $\video$ is processed by a student network $\studentNetwork$ on a task $\task$ (\eg, semantic segmentation for autonomous driving) and produces predictions $\prediction_i$ for each frame of the video $\image_i$ at the original video rate $\rate_\video$ (\ie, in real time). In parallel in the slow route (training), a frozen teacher $\teacherNetwork$ produces pseudo ground-truths $\pseudogroundtruth_{i'}$ from a subset of frames $\image_{i'}$ at a slower rate $\rate_\teacherNetwork$. The pair $(\image_{i'},\pseudogroundtruth_{i'})$ are then stored in an online dataset (or replay buffer) $\onlineDataset$ through an update function $\updateFunction$. $\onlineDataset$ is sampled through a selection function $\selectionFunction$ and the selected pairs ${(\image_{n},\pseudogroundtruth_{n})}$ are used to train a copy of the student network $\studentNetworkCopy$ for one epoch using a loss $\mathcal{L}$. The parameters $\weight$ of $\studentNetworkCopy$ are then transferred to $\studentNetwork$ at a rate $\rate_{\studentNetworkCopy}$ (corresponding to the inverse of the training time of $\studentNetworkCopy$ on one epoch) so that $\studentNetwork$ improves on the latest domain of $\video$. One of the contribution of our paper consists in including replay-based Continual Learning (CL) methods, $CL_{Rep}$, inside $\onlineDataset$ and regularization-based methods, $CL_{Reg}$, on $\loss$. }
    \label{fig:pipeline}
\end{figure*}

In this section, we first describe online distillation in a mathematical framework suited for continual learning. Next we detail the regularization-based and replay-based continual learning methods that we integrate into the online distillation framework. Finally, we explain how to evaluate and benchmark online continual leaning methods under our cyclic stream.

\subsection{Online distillation framework}
The online distillation framework proposed by Cioppa~\etal~\cite{Cioppa2019ARTHuS} allows a real-time network to adapt to domain shifts at test time.
Formally, given a long untrimmed video $\video$ composed of a stream of frames $\image_i$ produced at a rate $\rate_\video$ and a task $\task$ (\eg, object detection, semantic segmentation, \emph{etc}.), the objective is to produce a stream of predictions $\prediction_i$ for each frame $\image_i$ in real time (\ie, at a rate $\rate_\video$). To do so, the authors leverage a student-teacher architecture with a fast and slow route. In the fast route (inference), a student network $\studentNetwork$ computes $\prediction_i = \studentNetwork(\image_i)$ at the rate $\rate_\video$. 
In  parallel in the slow route (training), a slower but high-performance frozen teacher network $\teacherNetwork$ produces pseudo ground-truths $\pseudogroundtruth_{i'} = \teacherNetwork(\image_{i'})$ at an asynchronous slower rate $\rate_\teacherNetwork$ on a subset of $\video$.
Each new pair $(\image_{i'}, \pseudogroundtruth_{i'})$ is then stored through an update function $\updateFunction$ into an online dataset $\onlineDataset$ of size $\ODSize$ that is used to train a copy $\studentNetworkCopy$ of the student network. In the original framework, $\updateFunction$ is chosen as a First In First Out (FIFO) algorithm. 
Iteratively, $\studentNetworkCopy$ is trained on selected samples extracted from $\onlineDataset$ by a function $\selectionFunction$, by minimizing the loss: $$\loss = \sum_{n=1}^{N}\distance(\studentNetworkCopy(\image_{n}), \pseudogroundtruth_{n}) \ ,$$
where $\distance$ is a distance function suited to learn task $\task$. In the original framework, $\selectionFunction$ selects all pairs in $\onlineDataset$ one time.
The parameters of $\studentNetwork$ are updated by copying the parameters $\weight$ of $\studentNetworkCopy$ at the rate $\rate_{\studentNetworkCopy}$, corresponding to the inverse of the training time of $\studentNetworkCopy$ on one epoch of $\onlineDataset$. The complete pipeline may be found in Figure~\ref{fig:pipeline}.

Thanks to this framework, $\studentNetwork$ becomes specialized to the last minutes of the particular video it is analyzing. 
This allows it to adapt to slowly changing domains in $\video$ as long as $\teacherNetwork$ is able to produce reliable predictions.
However, this continual fine-tuning makes it forget previously acquired knowledge over time. For instance, when sudden shifts in domain occurs, $\studentNetwork$ needs several updates to recover good performance even if the same domain already appeared in the video.
In the following, we propose to incorporate Continual Learning (CL) techniques in the existing online distillation framework to minimize the catastrophic forgetting of previously acquired knowledge in the case of cyclic domain shifts. In particular, we benchmark several replay-based methods ($CL_{Rep}$) that act on $\onlineDataset$ and regularization-based methods ($CL_{Reg}$) that act on $\loss$ as shown in Figure~\ref{fig:pipeline}.


\subsection{Replay-based methods}
This set of methods leverage a replay buffer (\ie a collection of data and corresponding ground-truth labels) of finite size that is accessed by the selection function $\selectionFunction$ and updated with new data by an update function $\updateFunction$ at each training epoch.
The online distillation framework presented above can be formulated as a replay-based method, where the replay buffer corresponds to $\onlineDataset$, the labels are the pseudo ground-truth predictions $\pseudogroundtruth_{n}$, $\selectionFunction$ selects all data of the replay buffer to be used during the training epoch, and $\updateFunction$ determines the policy to update samples in the replay buffer. In the original online distillation framework, the size of the replay buffer is also the number of samples, $\ODSize$, passed to the model at each training step. 
We extend the replay buffer to include $\RBSize \geq \ODSize$ samples where we sample $\ODSize$ samples without replacement from the buffer at each training step.
We augment the selected samples with the new incoming data from the stream.



We consider several strategies to modify $\updateFunction$ and $\selectionFunction$ to reduce the catastrophic forgetting: FIFO, Uniform, Prioritized, and MIR.



\vspace{2pt}\noindent\textbf{FIFO}: $\updateFunction$ stores the most recent samples in the replay buffer while removing oldest ones. 
This is equivalent to the original framework's update strategy that is used as a baseline for comparison with other methods. 

\vspace{2pt}\noindent\textbf{Uniform}: $\updateFunction$ stores incoming data at randomly selected replay buffer indices. This strategy leads to an expected remaining lifespan of data to decay exponentially~\cite{Barnich2011ViBe}, which could avoid forgetting. As for memory selection $\selectionFunction$, it performs a random selection from memory for constructing a training batch.

\vspace{2pt}\noindent\textbf{Prioritized}: Adapting the work of Schaul~\etal~\cite{Schaul2015Prioritized-arxiv} on reinforcement learning, we set $\updateFunction$ to assign an importance score $\importance$ for each sample in the replay buffer following:
\[\importance_{n} = \distance(\studentNetwork(\image_{n}), \teacherNetwork(x_{n})) \label{Importance_score}\, .\]
The importance score is then used as a probability of determining which samples to remove from the replay buffer following:
$$\probability_{n} = \frac{\importance_{n}^{-1}}{\sum_{n'=1}^{\RBSize}\importance_{n'}^{-1}} \ .$$
To perform the memory selection $\selectionFunction$ operation, prioritized follows the same strategy described above
for the update function $\updateFunction$.

\vspace{2pt}\noindent\textbf{MIR~\cite{aljundi2019online}}: is a selection function $\selectionFunction$ that selects a subset of the replay buffer samples that are maximally interfered by the incoming data in a stream. In other words, it constructs a set of training samples from memory that are negatively affected the most by the next parameter update.

\subsection{Regularization-based methods}
Regularization-based methods mitigate forgetting by adding a regularization term to the training loss function $\loss$. Generally, this can be formulated as:

$$\loss = \sum_{n=1}^{N}\distance(\studentNetworkCopy(\image_{n}), \pseudogroundtruth_{n}) + \regularisation\ ,$$

\noindent where $\regularisation$ is a method-specific regularization term. In this paper, we consider four different regularization-based continual learning methods: ER-ACE~\cite{caccia2022new}, LwF~\cite{li2017learning}, MAS~\cite{aljundi2018memory}, and RWalk~\cite{chaudhry2018riemannian}. We summarize these methods hereafter.

\vspace{2pt}\noindent\textbf{ER-ACE~\cite{caccia2022new}} aims at reducing the changes in the learned representation when training on samples from a new class. It does so by applying an asymmetric parameter update on the incoming data and the previously seen data that are sampled from a replay buffer. Specifically, ER-ACE restricts the loss computation on  classes presented in the incoming data while ignoring remaining classes. We note that ER-ACE only works on incoming data while keeping the original loss on the data sampled from replay buffer.

The following methods were originally proposed for settings with clear task boundaries. 
We adopt them to work on online streams without task boundaries by using two properties: \textbf{(i)} warmup and \textbf{(ii)} update frequency. 
The warmup defines a time period for the network to be initialized during the warmup phase, we set $\mathcal R=0$.
The update frequency simulates an artificial task boundary after every $\updateFreq$ steps, where $\updateFreq$ is a fixed hyperparameter for all methods. 

\vspace{2pt}\noindent\textbf{LwF~\cite{li2017learning}} uses knowledge distillation to encourage the current network's output to resemble that of a network trained on data from previous time steps. In our setup, LwF keeps a previous version of our student network $\studentNetworkCopy$ to guide the future parameter updates of this network. Maintaining a previous network that is potentially more tailored to previous domains could help in preserving learned knowledge.

\vspace{2pt}\noindent\textbf{MAS~\cite{aljundi2018memory}} assigns an importance weight for each network parameter by approximating the sensitivity of the network output to a parameter change. When training on new distributions, it penalizes large changes to important parameters and, thus, preserves previously learned knowledge. 

\vspace{2pt}\noindent\textbf{RWalk~\cite{chaudhry2018riemannian}} is a generalized formulation that combines a modified version of the two popular importance-based methods: EWC~\cite{kirkpatrick2017overcoming} and PI~\cite{zenke2017continual}. 
RWalk computes importance scores for network parameters, similar to MAS, and regularizes over the network parameters.

\subsection{Evaluation methodology}
To evaluate the adaption to new domains and the forgetting of past domains, we propose several evaluation metrics. 
Following the work of Cioppa~\etal~\cite{Cioppa2019ARTHuS}, the performance of the student network $\studentNetworkCopy$ (equivalent to $\studentNetwork$) over time is defined as follows: 
given a task-specific metric $\metric$ (\eg, $\metricsegmentation$ for semantic segmentation or $accuracy$ for classification), a set of size $\evaluationsize$ of frames $\imageset_i'=\{\image_{i'}, ..., \image_{i'+\evaluationsize}\}$ and pseudo ground truths $\pseudogroundtruthset_i' =\{\pseudogroundtruth_{i'}, ..., \pseudogroundtruth_{i'+\evaluationsize}\}$, the performance of the student network at time $i'$ is given by:
$$\metric(\studentNetworkCopy(\imageset_{i'};\weight_{i'}), \pseudogroundtruthset_{i'}) \ , $$
where $\weight_{i'}$ are the parameters of $\studentNetworkCopy$ at time $i'$, which may be asynchronous with the training of $\studentNetworkCopy$ and update of $\studentNetwork$ as it operates at a the different rate $\rate_{\studentNetworkCopy}$.



\vspace{2pt}\noindent\textbf{Backward Transfer (BWT)}: Motivated by the discrete implementation of backward transfer~\cite{diaz2018don}, we propose a modified version for online streams that measures forgetting of the current student network with respect to previous data, which corresponds to the previous domain in our case:  
$$\text{BWT}(i') = \metric(\studentNetworkCopy(\imageset_{i'-h};\weight_{i'}), \pseudogroundtruthset_{i'-h}) \ , $$ 
where $h$ refers to the backward time shift. 

In addition, we report the \textbf{Final Backward Transfer (Final BWT)}. Given a stream of length $K$, we evaluate the backward transfer of the final model $\theta_K$ on the entire stream, \ie setting $h=0$ in BWT.
This metrics allows to evaluate the final student model on all previous domains, rather than only one specific past domain.



\vspace{2pt}\noindent\textbf{Forward Transfer (FWT)}: Similar to the backward transfer, we adapt the discrete version~\cite{diaz2018don} of forward transfer for our online setup as follows: 

$$\text{FWT}(i') = \metric(\studentNetworkCopy(\imageset_{i'+h};\weight_{i'}), \pseudogroundtruthset_{i'+h}) \ . $$

\vspace{2pt}\noindent Forward transfer measures the model's performance on future unseen data. In our case, this metric is useful in evaluating the current model on the next domain.
\section{Experiments}
\label{sec:experiments}

In this section, we first describe the experimental setup on which we benchmark our continual online distillation framework. Next, we provide quantitative results including a comparative study, of our framework using our proposed evaluation methodology. Finally, we display some qualitative results to show the practical impact for autonomous driving applications.
\begin{table*}[!ht]
\caption{\textbf{Quantitative results.} We compare several memoryless and replay-based methods with the original baseline framework proposed by Cioppa~\etal~\cite{Cioppa2019ARTHuS}. For each category, we benchmark several selection functions $\selectionFunction$, update functions $\updateFunction$, and regularizers $\regularisation$. The performance is provided for our proposed evaluation metrics for the \underline{20/40} concatenated sequences. The replay-based methods generally outperform the baseline and the memoryless methods. The LwF and MAS regularization methods decrease the performance, while ACE and RWalk increase the performance. The best results are obtained with a uniform replay buffer, MIR, MIR+ACE, and MIR+RWalk. We compare the temporal evolution of the performance of the \textcolor{blue}{Baseline} with one of the best performing method \textcolor{darkRed}{MIR+RWalk} in Figure~\ref{fig:quantitative}.}
    \centering
\begin{tabular}{l|lll||ccccc}
\toprule
 \multirow{2}{*}{\textbf{Methods}} & \multicolumn{3}{c||}{\textbf{Parameters}} & \multicolumn{5}{c}{\textbf{Metrics (mean \%)}} \\ 
{} & $\selectionFunction$ &   $\updateFunction$ & $\regularisation$ &  $\metricsegmentation$ &  $\metricsegmentation$ NDS &       FWT &       BWT &  Final BWT \\
\midrule
\multirow{4}{*}{Memoryless} & / & /  &  / & 18.4/19.4 &                 14.9/15.1 &  6.8/4.8 &  7.8/7.5 &   14.9/15.0 \\
 & / & / & MAS &  14.0/14.0 &                 13.0/13.3 &  11.1/11.1 &  12.9/12.9 &   14.2/14.2 \\
 & / & / & LwF &  15.7/15.9 &                 12.0/11.0 &  9.7/6.8 &  11.3/8.9 &   14.7/12.9 \\
 & / & / & RWalk &  18.3/19.3 &                 14.6/14.7 &  7.5/4.7 &  8.6/6.5 &   15.1/14.2 \\ \midrule \rowcolor[HTML]{f3f2fd}
 Baseline  & All & FIFO & /          &  23.4/24.2 &                 19.8/18.2 &  14.5/9.5 &  17.7/13.9 &   21.9/19.9 \\ \midrule
\multirow{7}{*}{Replay Buffer} & Uniform  & Uniform & /             &  25.5/25.0 &                 23.6/21.1 &  \textbf{22.2}/17.3 &  30.6/28.8 &   29.4/28.4 \\
 & Prioritized & Prioritized & /             &  25.1/25.1 &                 23.2/20.8 &  21.3/17.3 &  29.2/28.4 &   29.2/28.9 \\
 & MIR      & Uniform & /        &  25.2/25.2 &                 23.7/\textbf{24.5} &  21.9/\textbf{22.5} &  30.5/28.6 &   29.5/29.7 \\
 & MIR & Uniform & MAS              &  14.5/14.9 &                 13.4/14.7 &  12.1/13.6 &  13.9/15.2 &   15.1/15.4 \\
 & MIR & Uniform & LwF              &  18.7/18.1 &                 17.6/15.7 &  17.4/13.9 &  21.0/20.2 &   22.4/21.1 \\
  & MIR & Uniform & ACE              &  \textbf{25.6}/\textbf{25.5} &                 \textbf{24.2}/21.8 &  22.0/17.5 &  \textbf{30.8}/29.4 &   28.8/28.5 \\ \rowcolor[HTML]{fdf2f3}
 & MIR & Uniform & RWalk            &  25.2/25.4 &                 23.4/22.0 &  21.8/18.0 &  30.0/\textbf{30.8} &   \textbf{30.1}/\textbf{30.8} \\
\midrule
\bottomrule
\end{tabular}
    \label{tab:quantitative}
\end{table*}

\subsection{Experimental setup}

Our online continual learning framework is agnostic to the task, metric, and training parameters. In this section, we provide the technical details describing our experiments in various settings.

\mysection{Task.}
We benchmark our framework on the outdoor semantic segmentation task, which consists in assigning a class label to each pixel of a frame. We study the particular case of videos taken behind the windshield of vehicles, which is the typical study-case for autonomous driving applications.

\mysection{Dataset.}
The online distillation framework requires long untrimmed videos, in our case containing cyclic domain shifts. Additionally, these videos must be relevant to highlight the task's objectives. Since most datasets for semantic segmentation are composed of frames or small video clips (\eg, CityScapes~\cite{Cordts2016The}, BDD100K~\cite{Yu2020BDD100K}, \etc), they cannot be used in our context of online continual learning.
Hence, we follow the same strategy to simulate long videos with domain shifts as in \cite{Cioppa2019ARTHuS} and  propose to artificially construct a video $\video$ by concatenating sequences from $2$ different domains, $\domain{A}$ and $\domain{B}$, alternating in cycle from one domain to the other. The resulting video is therefore an ordered set $\video = \{\videodomain{A}_1, \videodomain{B}_1, \videodomain{A}_2, \videodomain{B}_2, ...\}$, where the $\videodomain{A}_i$ and $\videodomain{B}_i$ are sequences from domain $\domain{A}$ and domain $\domain{B}$, respectively. 
In our autonomous driving case, we define the two domains $\domain{A}$ and $\domain{B}$ as a highway environment and a downtown environment, which differ from the priors on the semantic classes (\eg, there should be fewer persons in highways than downtown) or the background (\eg, there are more buildings in downtown and more empty spaces in highways). We extract several clips from each domain and alternatively concatenate them to build $\video$. To consider clips of different time lengths, we construct two video $\video$ streams where the extracted clips are 20 minutes and 40 minutes long respectively.


\begin{figure*}[!ht]
    \centering
    \begin{subfigure}{0.49\textwidth}
        \centering
        \includegraphics[width=\textwidth]{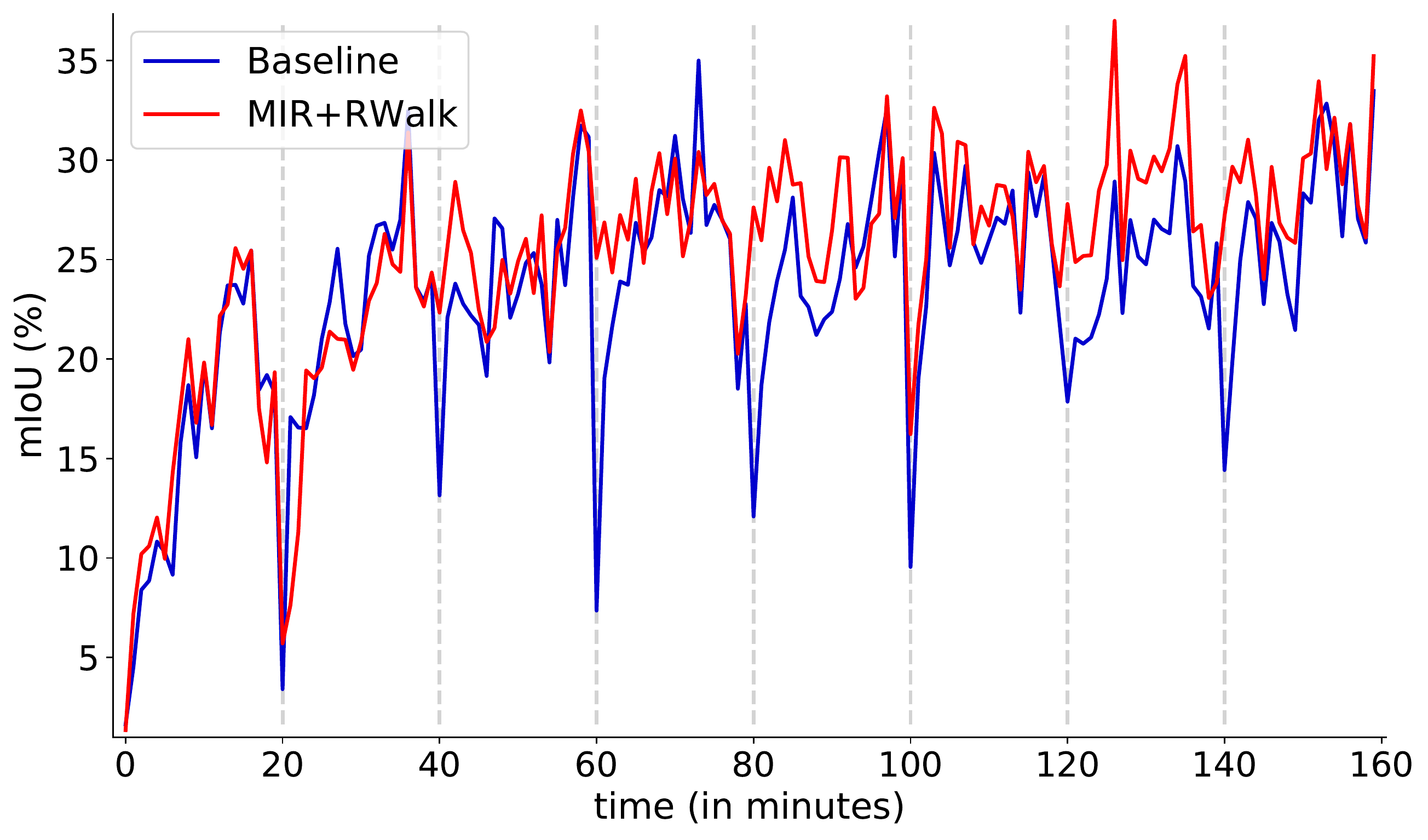}
    \end{subfigure}
    \begin{subfigure}{0.49\textwidth}
        \centering
\includegraphics[width=\textwidth]{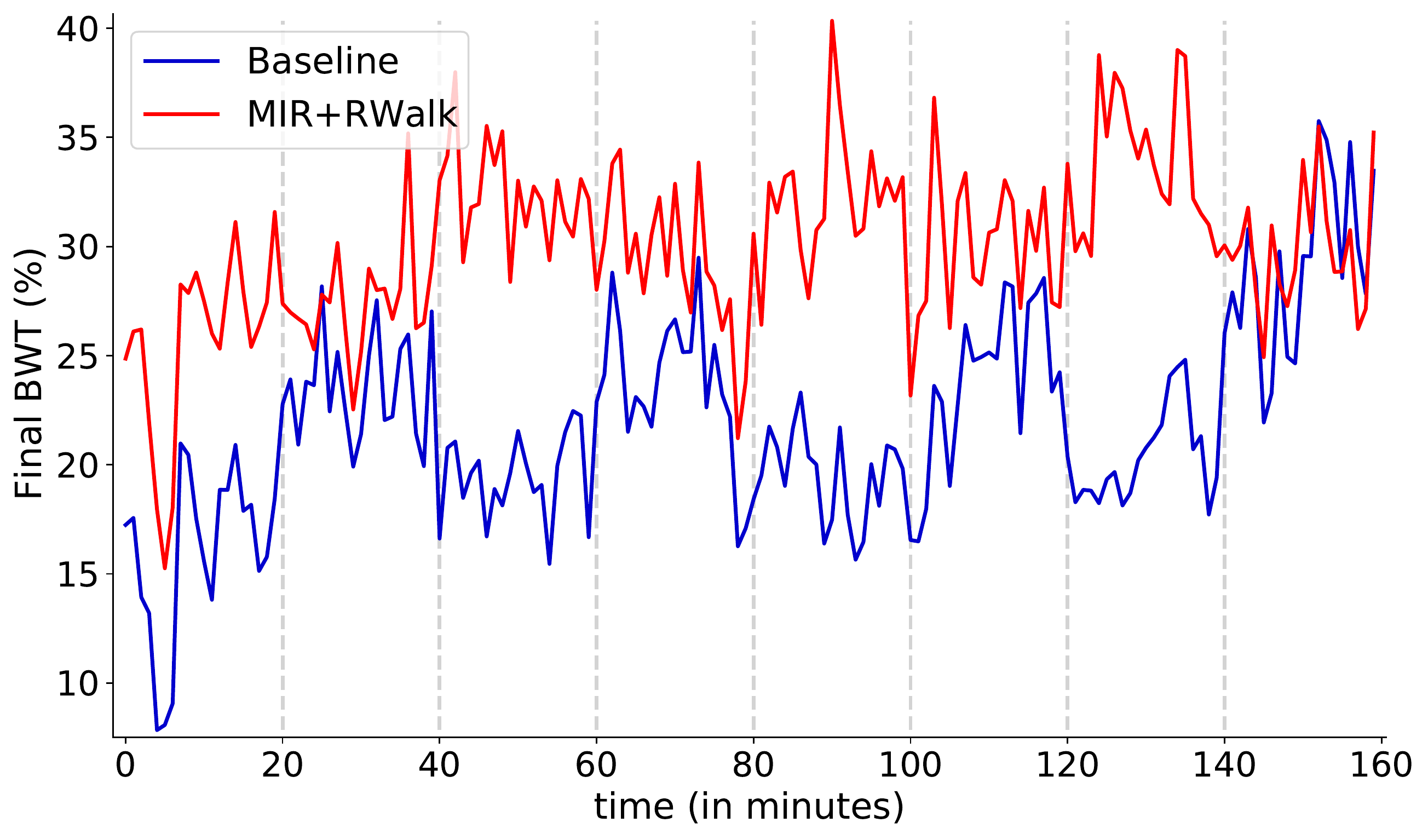}
    \end{subfigure}
    \begin{subfigure}{0.49\textwidth}
        \centering
\includegraphics[width=\textwidth]{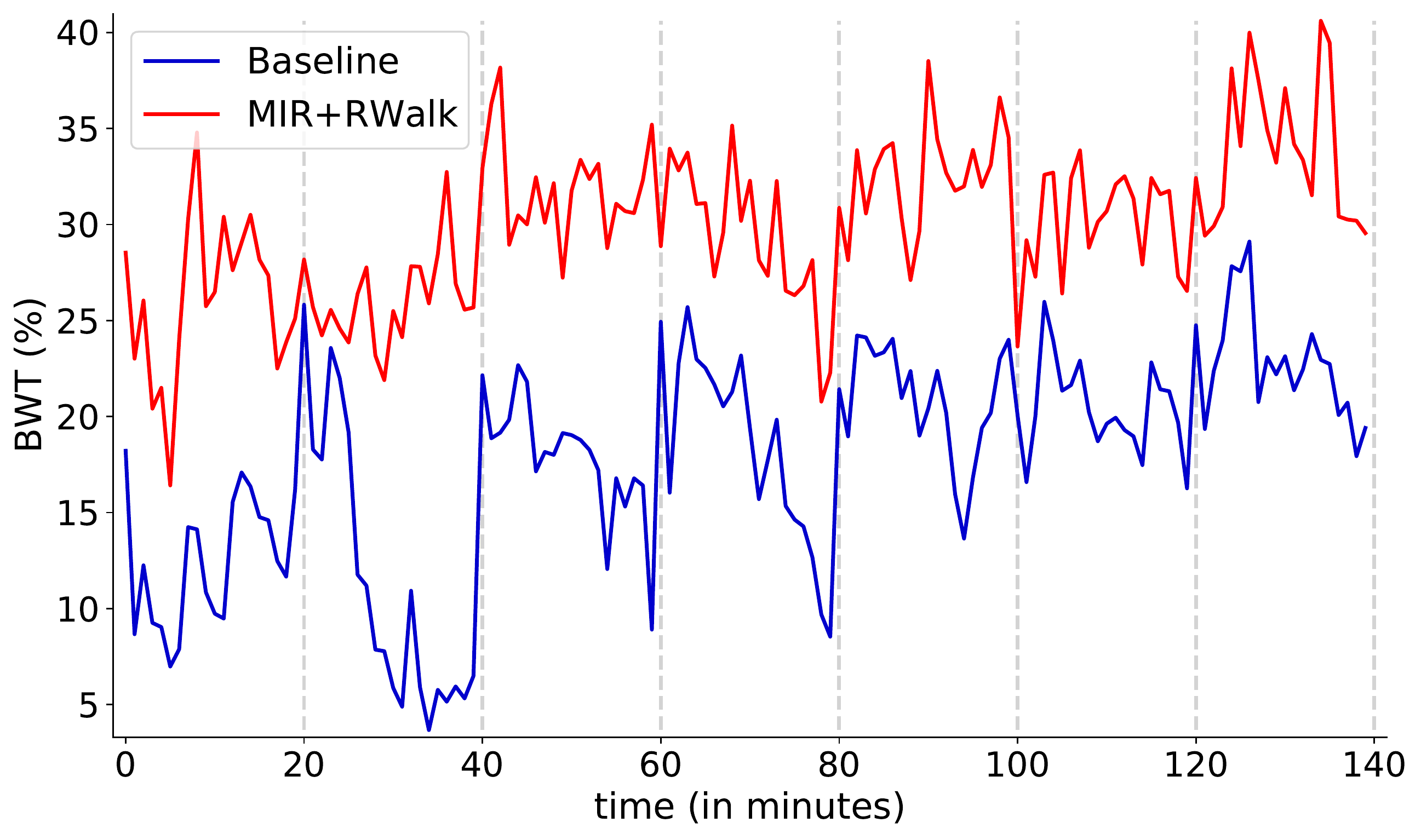}
    \end{subfigure}
    \begin{subfigure}{0.49\textwidth}
        \centering
\includegraphics[width=\textwidth]{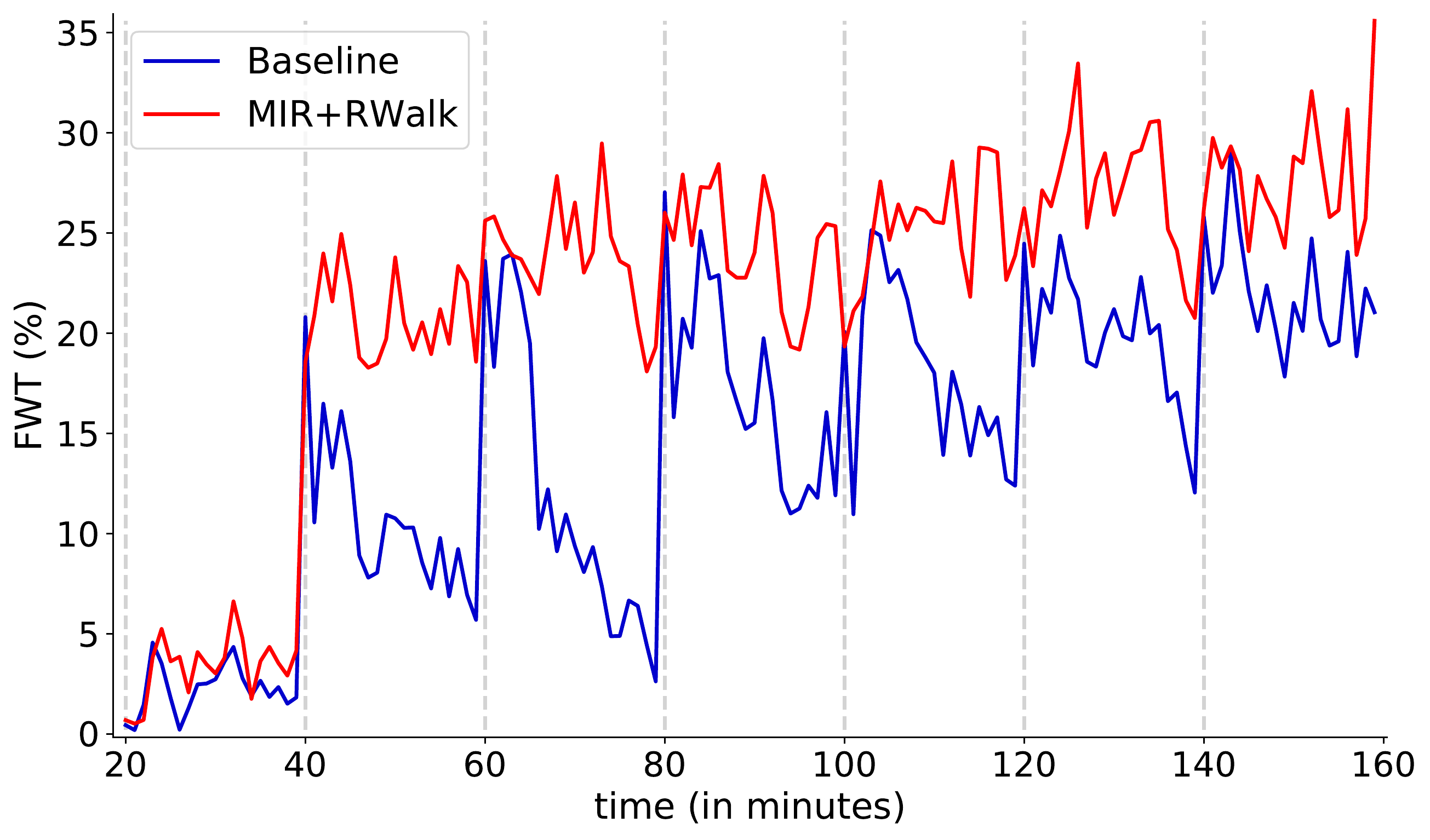}
    \end{subfigure}
    \caption{\textbf{Evolution of the performance over time.} We compare the evolution with respect to $\metricsegmentation$, BWT, Final-BWT, and FWT of the MIR+RWalk method with the original online distillation framework (baseline). (Top-left) $\metricsegmentation$: the performances are mostly similar within the domain, but around the domain shifts (from the second cycle), the baseline suffers from forgetting while MIR+RWalk keeps high performance. (Bottom-left) BWT: when evaluating on the previous domain, MIR+RWalk clearly outperforms the baseline, showing that it is able to retain information about the previous domain, on frames it has trained on. (Top-right) Final-BWT: the baseline quickly forgets past knowledge, while MIR+RWalk is able to retain high performance for both domains across many cycles. (Bottom-right) FWT: when evaluating on the future domain, MIR+RWalk also significantly outperforms the baseline, showing that it is able to generalize on new frames of a particular domain using information from a previous domain it has seen before. }
    \label{fig:quantitative}
\end{figure*}

\mysection{Evaluation metric.}
Following the standards in semantic segmentation, we use $\metric = \metricsegmentation$ to evaluate the segmentation masks of each frame as described in Section~\ref{sec:method}. Following the work of Cioppa~\etal~\cite{Cioppa2019ARTHuS}, since ground-truth data is unavailable for our dataset, we evaluate the performance of the student with respect to the pseudo ground truths produced by the teacher. This evaluates the capacity of the student to imitate the teacher.
We provide the $\metricsegmentation$, FWT, BWT, Final BWT metrics either during the video or averaged over the entire video (referred as mean). We choose $I=1$ minute and $h=20$ minutes or $h=40$ minutes depending on the domain sequences length to evaluate the forgetting on the previous or future domains.
Finally, we also compute the average across a time window of $\pm$ 2 minutes of each domain shift occurrence. We call this metric $\metricsegmentation$ Near Domain Shifts ($\metricsegmentation$ NDS). 

\mysection{Networks and training parameters.}
For the teacher network $\teacherNetwork$, we chose SegFormer~\cite{Xie2021SegFormer} trained on the CityScapes dataset, which is the state of the art in semantic segmentation on this dataset.
For the student networks $\studentNetwork$ and $\studentNetworkCopy$, we chose TinyNet~\cite{Cioppa2019ARTHuS,Cioppa2018ABottomUp}, a lightweight segmentation network that only needs a few training samples to specialize on a particular domain, that is fast to train, and operates in real time (at least $30$ frames per second for full-HD videos on a Nvidia 1080 GPU).
The student network $\studentNetworkCopy$ is trained from scratch at the beginning of the video using a learning rate of $10^{-4}$ and ADAM optimizer for online learning following~\cite{Cioppa2019ARTHuS}. 
The replay buffer size is set to $\RBSize=250$ and the number of selected frames to $\ODSize=100$ frames.   
Given the chosen video, networks, and replay buffer size, the rates are: $\rate_\video=30$ frames per second, $\rate_\teacherNetwork=3$ seconds per frame, and $\rate_{\studentNetworkCopy}=60$ seconds per epoch.

\newcommand\rowincludegraphics[2][]{\raisebox{-0.45\height}{\includegraphics[#1]{#2}}}

\begin{figure*}[!ht]
    \centering
    \begin{tabular}{c|c|c|c|c}
    \textbf{RGB Image} &\textbf{Ground truth} & \textbf{Baseline} & \textbf{MIR} & \textbf{MIR+RWalk} \\ 
       \rowincludegraphics[scale=0.06]{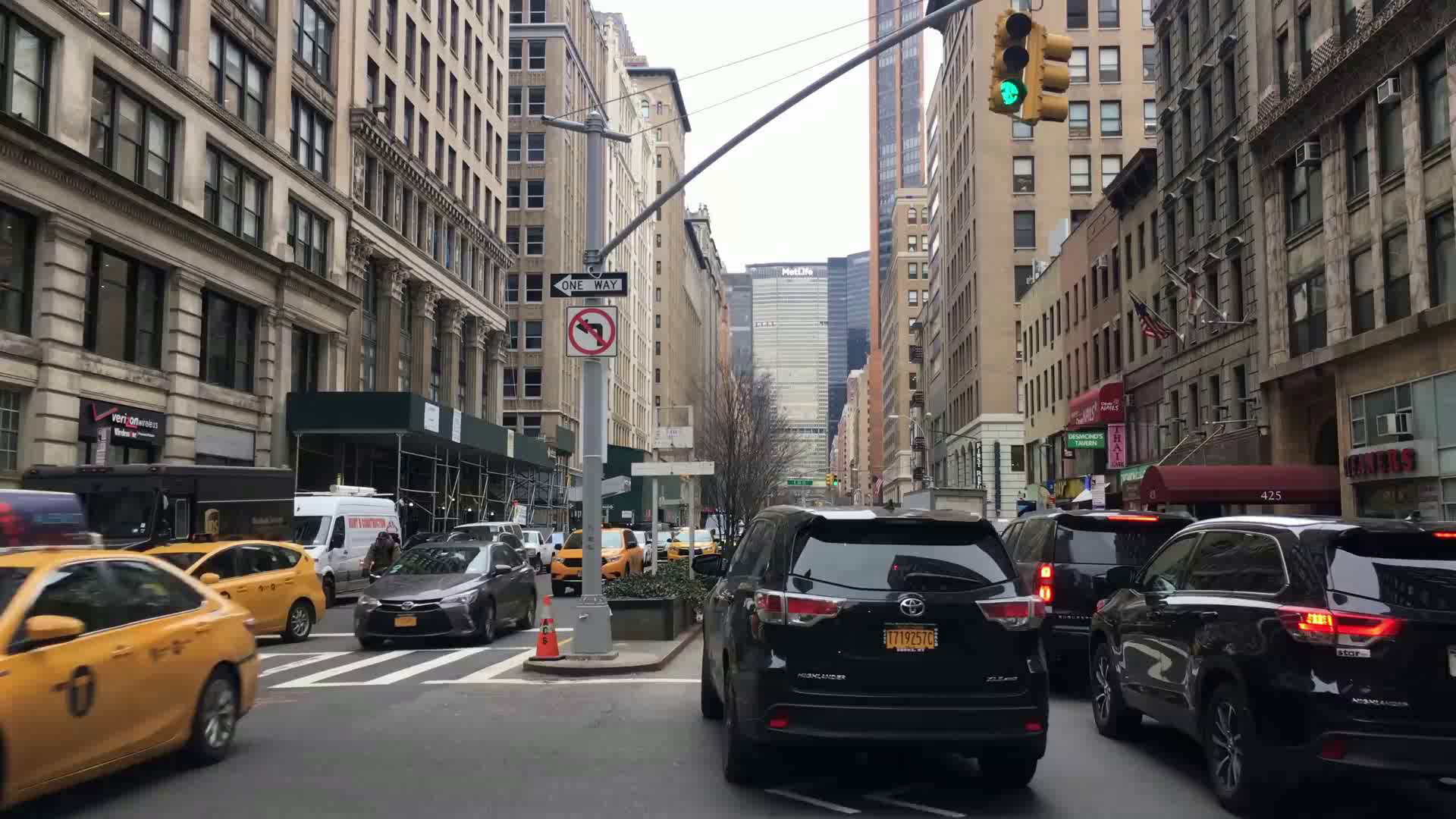} &
       \rowincludegraphics[scale=0.06]{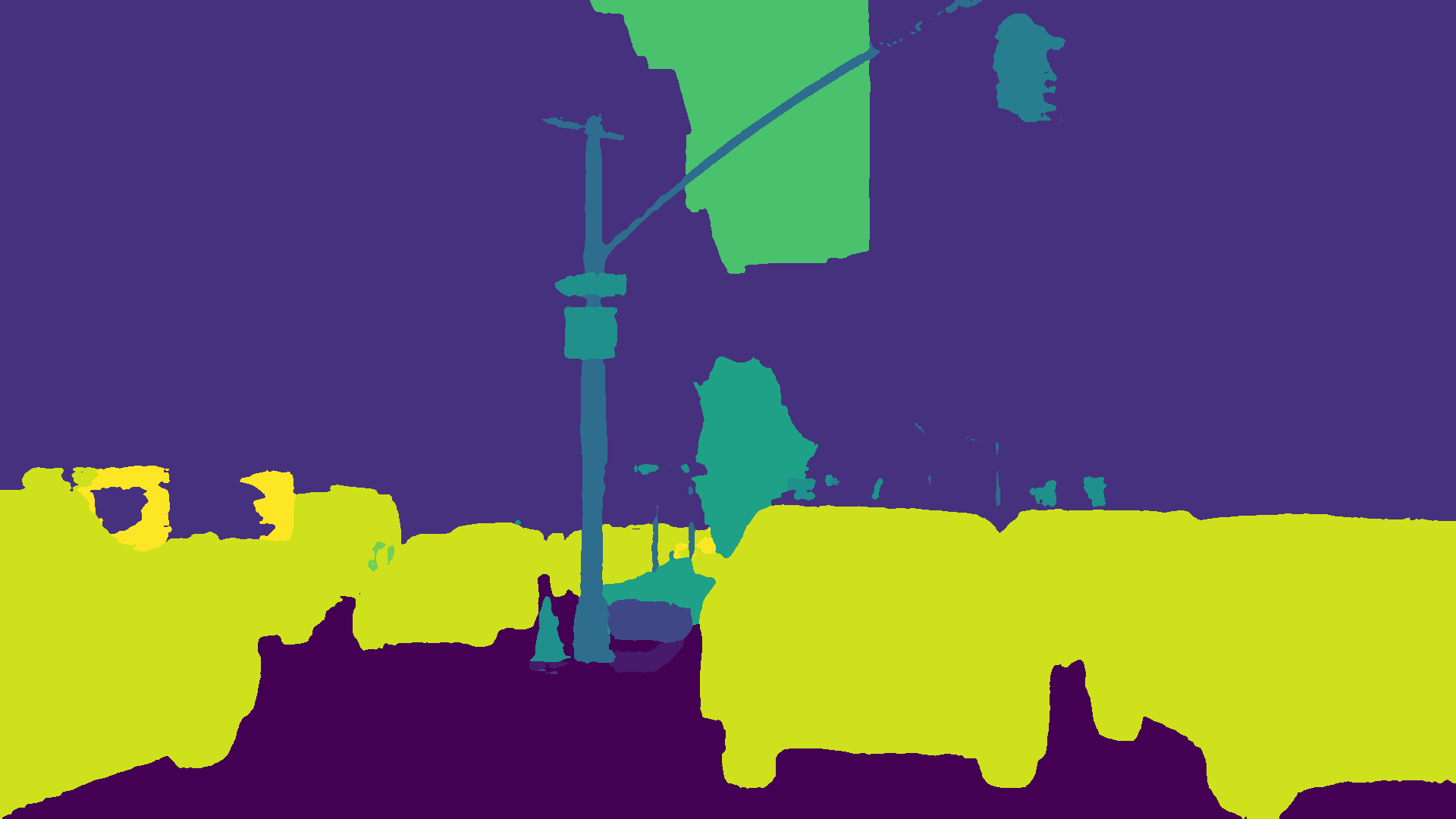} & \rowincludegraphics[scale=0.06]{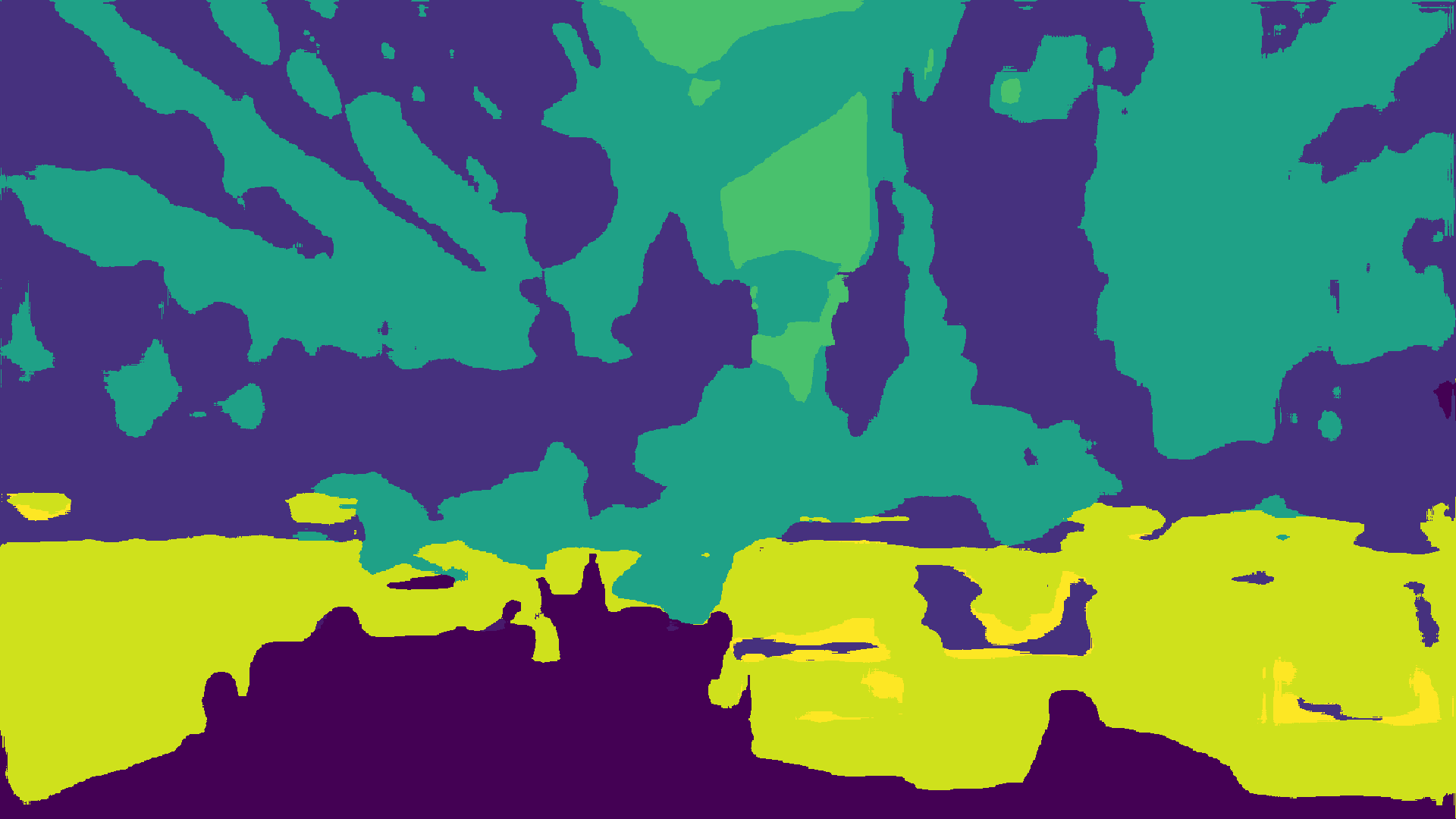} & \rowincludegraphics[scale=0.06]{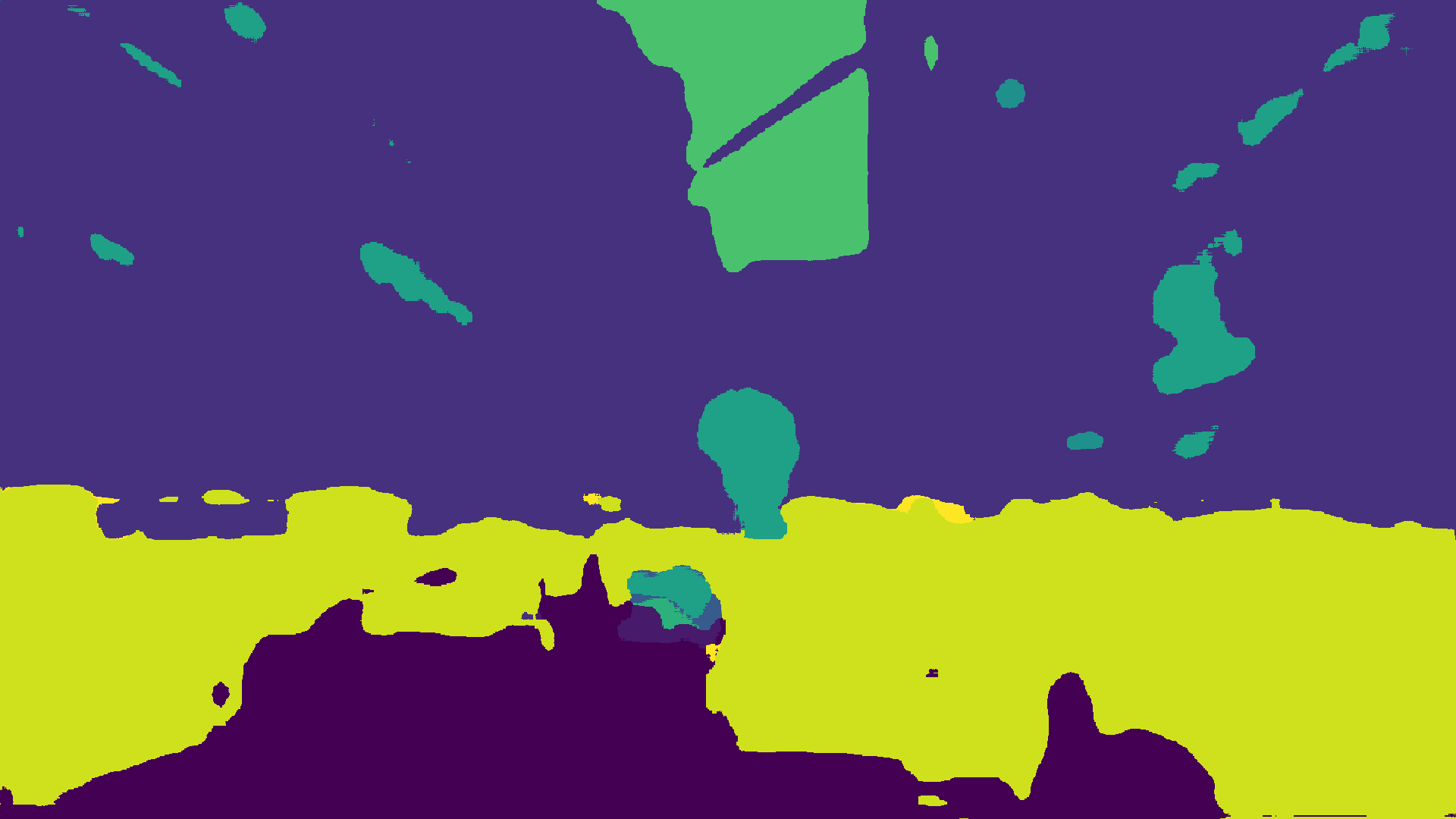} & \rowincludegraphics[scale=0.06]{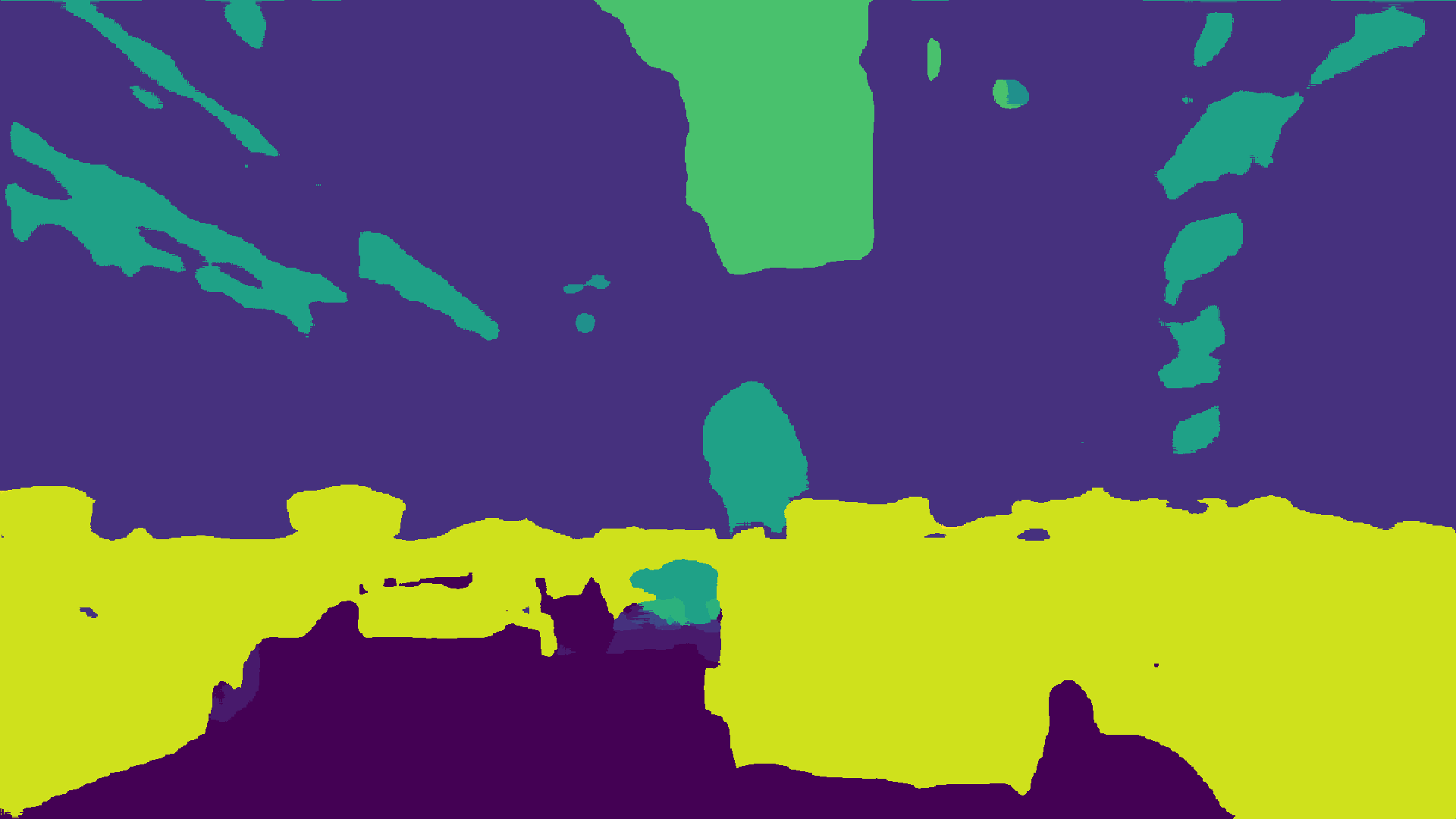} \\
       \rowincludegraphics[scale=0.06]{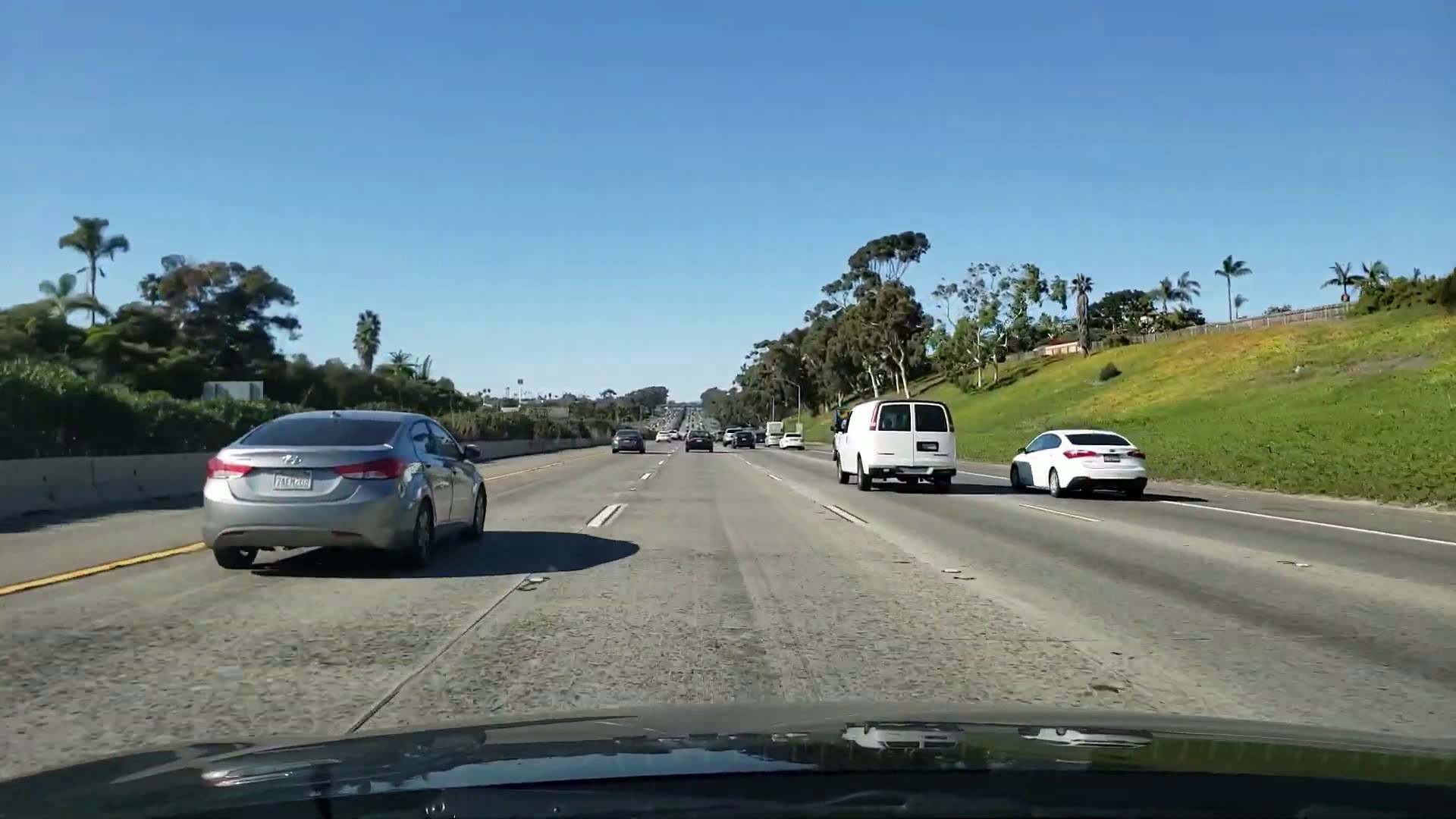} &
       \rowincludegraphics[scale=0.06]{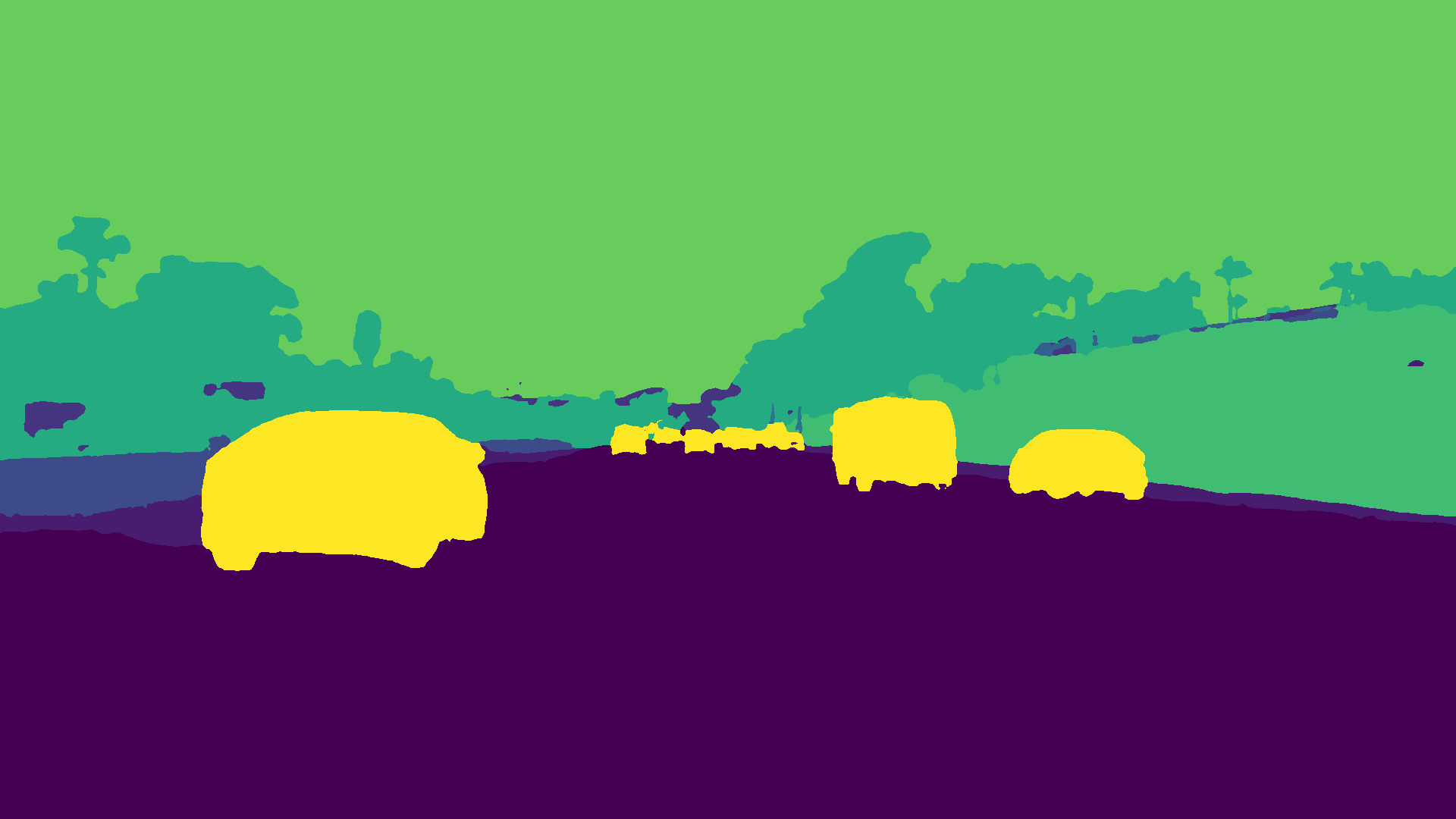} & \rowincludegraphics[scale=0.06]{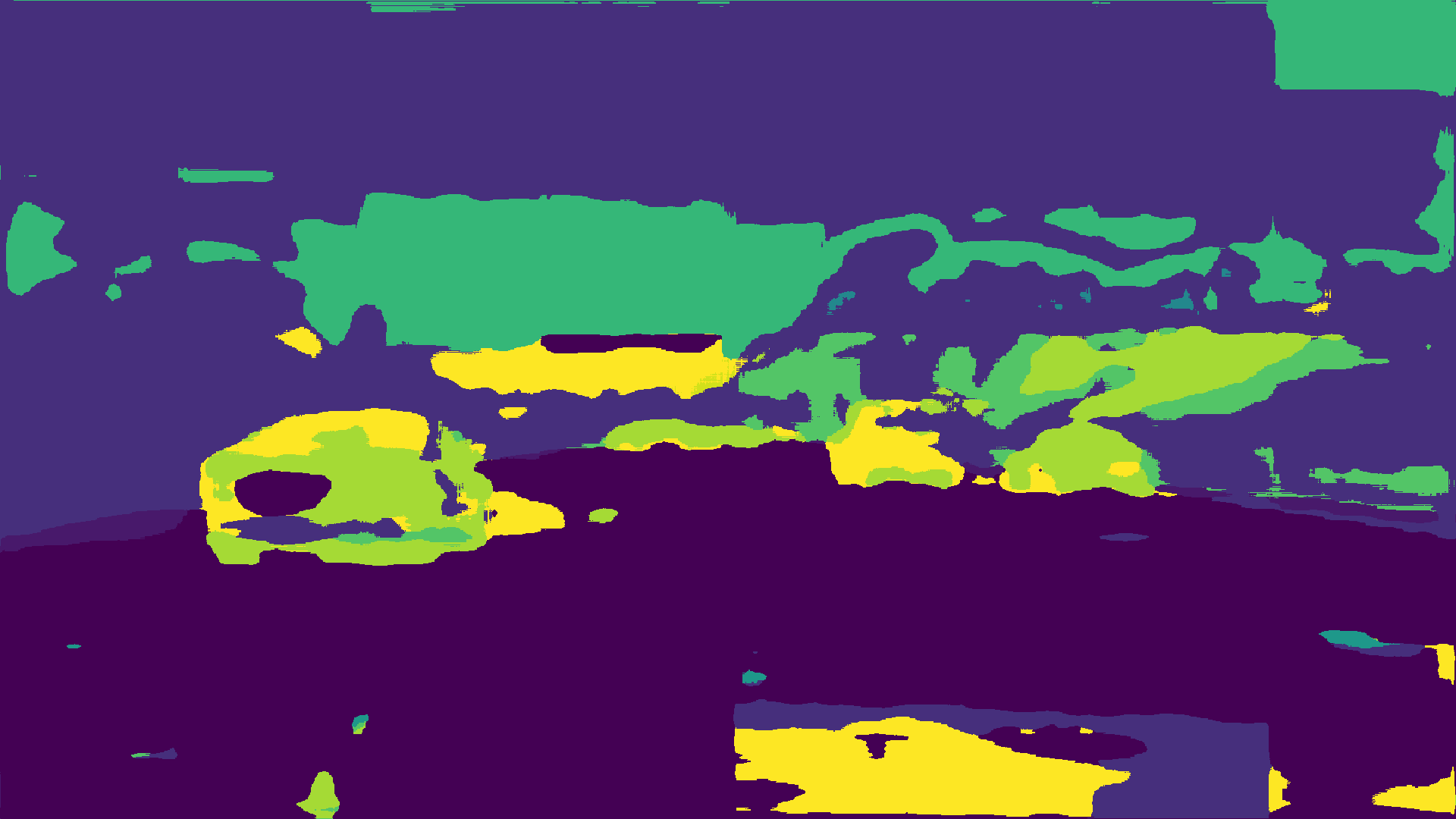} & \rowincludegraphics[scale=0.06]{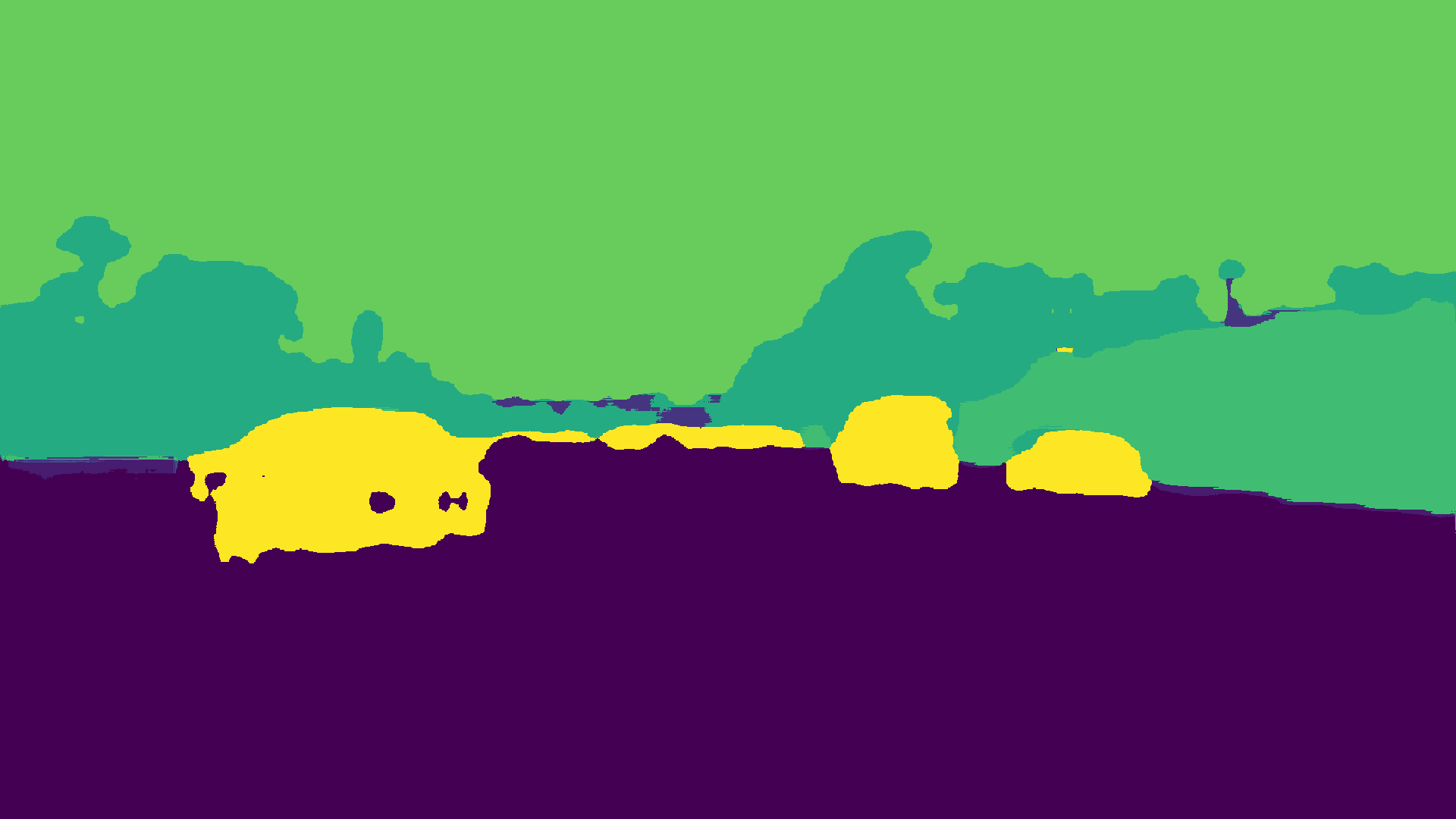} & \rowincludegraphics[scale=0.06]{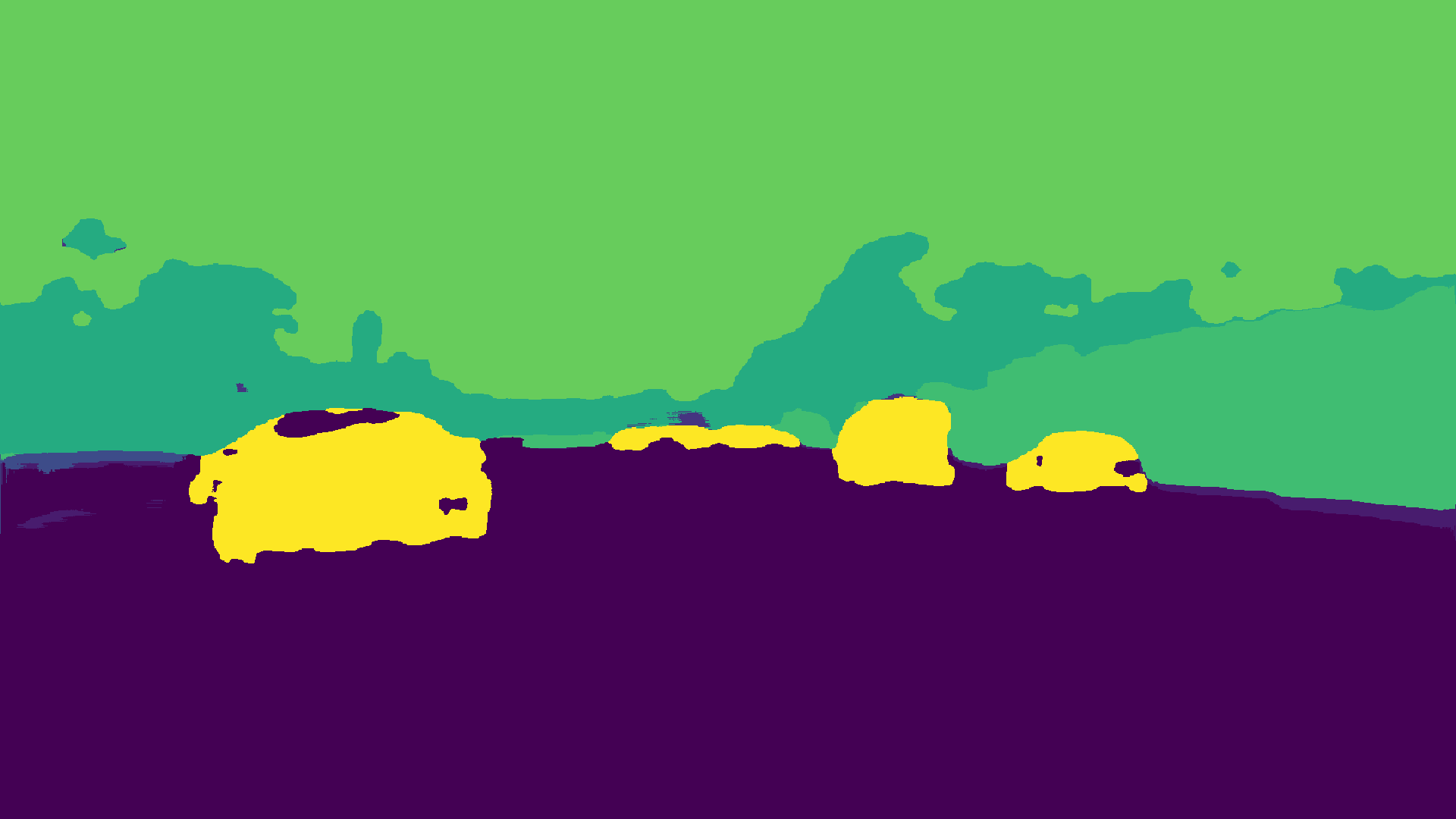} \\ 
    \end{tabular}
    \caption{\textbf{Qualitative results.} Comparison of the segmentation masks obtained by different online continual learning methods: (top row) a frame taken right after second transition between highway and downtown, and (bottom row) a frame taken right after seventh transition between downtown and highway. The baseline method predicts poor segmentation masks after the domain shift, even though it has already seen this domain before. In contrast, MIR and MIR+RWalk produce better segmentation masks.}
    \label{fig:qualitative}
\end{figure*}

\subsection{Quantitative results}

We compare the performance of the original framework with the proposed continual learning approaches.
As a naive approach, we also study a memoryless online distillation framework, in which the online dataset does not store any frame. In this setup, the pairs produced by the teacher are used only once for training and are then deleted.
As can be seen from Table~\ref{tab:quantitative}, the memoryless approaches perform worse than the original framework for all metrics, showing that retaining some information in an online dataset (or replay buffer) improves the performance.
Interestingly, all replay-based methods without regularizers improve compared to the baseline, with the best performance obtained by MIR overall. 
Adding a regularizer is however not always beneficial. For instance, MAS and LwF systematically decrease the performance, while ACE and RWalk slightly increase the performance.
We hypothesize that this can be attributed to the fact that MAS and LwF were proposed in the offline setup with the aim of reducing the elasticity of the model towards adapting to new information.
While this approach was proven to be useful in several scenarios, it could hinder the student from quickly adapting to new domains in the online setup.
The biggest improvement is therefore mainly due to the replay buffer method with MIR.

In Figure~\ref{fig:quantitative}, we show the evolution of the performance over time for the baseline and on one of the best method (MIR+RWalk) for cycles of $20$. As can be seen from the $\metricsegmentation$ plot, during the two first cycles, both methods have similar results. This is expected as they both discover the new domains. The first difference can be seen at the second transition, where the first domain is seen once again. The baseline method has a huge drop, while the continual learning method shows good performance. At each other transition, MIR+RWalk does not suffer from the drop in performance caused by the forgetting of the previous domain.
We conduct a comparison between the MIR+RWalk method and the original online distillation framework (baseline) by analyzing the performance evolution of the $\metricsegmentation$, BWT, Final-BWT, and FWT metrics. When evaluated on the previous domain, MIR+RWalk significantly outperforms the baseline in BWT, indicating its ability to retain information about the previous domain on frames it has been trained on. In the case of Final-BWT, the baseline quickly forgets past knowledge, while MIR+RWalk is able to maintain high performance for both domains across many cycles. Finally, when evaluated on the future domain, MIR+RWalk also shows significant performance improvements  compared to the baseline in FWT, indicating its ability to generalize on new frames from a previous domain.

\subsection{Qualitative results}

We qualitatively demonstrate the effect of the best performing continual learning methods on the catastrophic forgetting. To do so, we investigate the quality of the segmentation masks right after the second transition from highway to downtown (the student has seen the downtown only once before), and the seventh's transition from downtown to highway (the student has already seen the highway domain $6$ times before).
Figure~\ref{fig:qualitative} compares the segmentation masks obtained by the baseline method, MIR, and MIR+RWalk with the ground-truth mask.
As shown, even though the student has already seen the domain previously, the segmentation masks of the baseline right after the domain shift are very poor. In practice, this could lead to hazardous situations for the autonomous vehicle and its passengers. On the contrary, the segmentation masks obtained with MIR and MIR+RWalk are much closer to the ground-truth masks. The quantitative results demonstrate that incorporating continual learning algorithms into the online distillation framework considerably enhances the quality of the predictions, rendering it more viable for real-world applications.

\section{Conclusion}
\label{sec:conclusion}

In conclusion, the development of online distillation has brought new opportunities for adapting deep neural networks in real time, making them more suitable for practical applications such as autonomous driving. However, the issue of catastrophic forgetting when the domain shifts has been a major challenge in the implementation of this technique. In this paper, we proposed a novel solution to this issue by incorporating continual learning methods. Through our experimentation, we evaluated several state-of-the-art continual learning methods and demonstrated their effectiveness in reducing catastrophic forgetting. We also conducted a detailed analysis of our proposed solution in the case of cyclic domain shifts. The results highlight that our approach improves the robustness and accuracy of online distillation, making it a promising technique for real-world applications. This work represents a significant step forward in the field of online distillation and continual learning, with the potential to have a meaningful impact on various fields such as autonomous driving.

\mysection{Acknowledgement.}
A.~Cioppa is funded by the F.R.S.-FNRS.
 A.~Halin is funded by the Walloon region (Service Public de Wallonie Recherche, Belgium) under grant No.~2010235 (ARIAC by DIGITALWALLONIA4.AI). M.~Henry is funded by PIT MecaTech, Belgium, under grant No.~C8650 (ReconnAIssance).
 This work was partially supported by the King Abdullah University of Science and Technology (KAUST) Office of Sponsored Research (OSR) under Award No.~OSR-CRG2021-4648.

{\small
\bibliographystyle{ieee_fullname}
\bibliography{bib/abbreviation-short,bib/add-new-refs-here-do-not-touch-others-bibfiles,bib/dataset,bib/labo,bib/ok,bib/segmentation}
}

\end{document}